\documentclass{article}
\usepackage{spconf,amsmath,graphicx}
\usepackage{multirow}
\usepackage{amsmath, color}
\usepackage{cite}
\usepackage[T1]{fontenc}

\ninept
\title{Internal Language model estimation based adaptive language model fusion for domain adaptation}
\name{Rao Ma, Xiaobo Wu, Jin Qiu, Yanan Qin, Haihua Xu, Peihao Wu, Zejun Ma}
\address{ByteDance}

\begin{document}
%
\maketitle
\begin{abstract}
%
ASR model deployment environment is ever-changing, and the incoming speech can be switched across different domains during a session. This brings a challenge for effective domain adaptation when only target domain text data is available, and our objective is to obtain obviously improved performance on the target domain while the performance on the general domain is less undermined.
In this paper, we propose an adaptive LM fusion approach called  internal language model estimation based adaptive domain adaptation (ILME-ADA). 
To realize such an ILME-ADA, 
an  interpolated  log-likelihood score is calculated based on the maximum of the  scores from the internal LM and the external LM (ELM) respectively. 
We demonstrate the efficacy of the proposed ILME-ADA method with both RNN-T and LAS modeling frameworks employing neural network and n-gram LMs as ELMs respectively on two domain specific (target) test sets. 
The proposed method can achieve significantly better performance on the target test sets while it gets minimal performance degradation on the general test set, compared with both shallow and ILME-based LM fusion methods.  


\end{abstract}
%
\section{Introduction}
\label{sec:intro}

Among diverse ASR modeling frameworks, end-to-end (E2E) ASR systems that effectively combine all components into a unifying network are predominantly prevailing \cite{graves2014towards, chan2016listen,vaswani2017attention,chiu2018state, gulati2020conformer}.  Nowadays, E2E systems have been widely deployed in commercial areas~\cite{hoy2018alexa,sainath2020streaming,li2022language}.
Although such commercial ASR systems are trained on tens of thousands of transcribed speech data and achieve good performance in general, training data cannot feasibly cover all possible fields. Effective ASR customization (aka domain adaptation) has therefore become a major concern, which aims to optimize speech recognition performance for specific domains based on the requirements of clients. For example, a medical client would like to improve the recognition accuracy of medical terms, and meanwhile maintain a good  ASR performance in other, say general domains. 

A common way to adapt the ASR system is to label as much speech data as possible from the target domain and re-train the  ASR model \cite{bell2020adaptation}. However, this approach is laborious and expensive to adopt in practice. Compared to transcribing speech data, collecting large-scale text data from the target domain is easier to obtain. For example, several works \cite{rosenberg2019speech, joshi2022simple} proposed to collect text from the target domain and generate ``fake'' speech data using  text-to-speech (TTS) model. Such methods are subject to the availability of a well-performing TTS system,  and more importantly, it still requires to fine-tune the original ASR model. Besides, with more domain specific text data, language model (LM) fusion \cite{toshniwal2018comparison} is also a popular solution in industry, where clients can  train LMs locally and perform LM fusion during inference for domain adaptation. The benefit of such a kind of recipe is that the ASR model remains unchanged and can continue to serve different purposes. 

For LM fusion, several  methods have been proposed, including shallow fusion \cite{toshniwal2018comparison}, cold fusion~\cite{sriram2018cold}, and deep fusion~\cite{gulcehre2015using}. Among all these fusion methods, shallow fusion is one of the most effective methods and has been widely applied in both research and product areas \cite{toshniwal2018comparison}. Recently, researchers have come to revisit the shallow fusion method under the E2E ASR framework to reduce the impact of internal language models (ILMs). 
The motivation comes from the fact that the outputs of the E2E models are normally posterior-based variables that contain a language model component implicitly learned from the training transcripts. 
The first pioneer works in this regard are hybrid auto-regressive transducer~\cite{variani2020hybrid} and log-likelihood density ratio methods~\cite{mcdermott2019density}. 
More recently, a series of ILM estimation (ILME) methods ~\cite{meng2021internal, zeyer2021librispeech, zeineldeen2021investigating, liu2022internal} have been proposed and better recognition results are achieved.


Though both shallow and ILME-based LM fusions have achieved decent performance in domain adaptation (DA) work,
a major concern of DA is to  let the ASR system avoid overfitting to the target domain and  obviously degrading on general domains.
This is challenging since ASR deployment environment is ever-changing and we cannot assume if the incoming speech is always from a specific domain. 
A simple idea to solve this problem is to build a LM by employing text data from a wide range of domains. Recently, \cite{choudhury2022likelihood} proposed a likelihood ratio based method by utilizing a separate LM trained from the ASR transcripts that are also exploited to train the external LM (ELM) as general data set. Unfortunately, text data from the general domain is normally very large and hard to access due to proprietary reasons
for clients. Besides, 
Training an LM  with bigger data would be  computationally intensive and time-consuming on the client side. 
Therefore,  given only text data from a target domain, how to train a LM that is not only lightweight (for instance, n-gram versus bulky Neurl Network LM (NNLM)) but also 
has good performance on the target domain while the performance on the general domain has been less affected is yet to be solved.


In this work, we propose an adaptive LM fusion approach to domain adaptation using ILME method. With only text data from the target domain for external language modeling,   the proposed method can achieve better performance 
on the target domain test sets while it gets  least performance degradation  on  general domain test sets, compared with the shallow as well as ILME-based fusion methods.
We denote such an adaptive LM fusion as ILME-ADA. Essentially, for each incoming token during the inference process, we maximize the log-likelihood scores between an ILM  and the corresponding ELM. We verify the efficacy of the proposed method over different test sets under two popular E2E ASR frameworks, RNN-T and attention-based encoder-decoder (AED) ones, as well as 
with different ELMs, such as Neural Network (NN) and n-gram LMs, fusion methods respectively.
\section{Related Work}~\label{sec:related}
To obtain balanced recognition results on two different domains  with LM fusion at the same time, one can think of a simple method, namely, building an interpolated LM by employing two data sets respectively. Unfortunately, ASR source data is normally very big and not easy to come by, say, for proprietary issues, ILME-based LM fusion~\cite{meng2021internal} can solve the problem of target domain recognition once more target domain text data is available. However, the ASR performance is degraded on the original/general domain data at the same time. 

Recently, \cite{choudhury2022likelihood} proposed a likelihood ratio based LM fusion method for DA while the ASR performance on the general data set is not degraded. To the best of our knowledge, the idea is the closest to what is proposed in this work. 
However, the proposed ILME-ADA is significantly different from \cite{choudhury2022likelihood} in several aspects. 
First, they employ likelihood ratio method, while it is the ILME method that is proposed here.
Secondly, the ELM in \cite{choudhury2022likelihood} is interpolated with both data of general and target domains, here, the ELM is always trained with only target domain data.
Thirdly, they only verified the idea with the n-gram LM fusion method over the RNN-T ASR model. Here we demonstrate the efficacy of the proposed method 
with RNNLM and n-gram LM over both RNN-T  and LAS ASR frameworks respectively.

\section{Proposed method}~\label{sec:proposed}
\subsection{Internal Language Model Estimation}~\label{sec:ilme}
For End-to-End (E2E) speech recognition, the inference process can be formulated as follows:
\begin{equation}
  \hat{W} = \arg\max_{W}\log p_{\theta}(W|X)\label{eq:asr-infer}
\end{equation}
where $\theta$ is the parameters of the E2E models, $X$ is input acoustic features,  $p_\theta(W|X)$ is the joint posterior generated by the E2E ASR models, while $\hat{W}$ is the word sequence that has the highest joint posterior among candidates. 

If we have extra text data, we can build an external LM, when the LM score is fused during ASR inference, Eq.~\ref{eq:asr-infer} is then turned into:
\begin{equation}
    \hat{W} = \arg\max_{W}[\log p_{\theta}(W|X) + \log p_{\text{LM}}(W) ] \label{eq:lmf}
\end{equation}
where $p_{\text{LM}}(W)$ is the joint probability of a word sequence $W$, and it is estimated with the joint conditional probability.
However, Eq.~\ref{eq:lmf} is a little bit conflicted with conventional Bayes theory in Hybrid ASR, which is as follows:
\begin{equation}
    \hat{W} = \arg\max_{W}[\log p_{\theta}(X|W) +   \log p_{\text{LM}}(W) ] \label{eq:bayes}
\end{equation}
Estimating likelihood score $p_{\theta}(X|W)$ under E2E model circumstance is not straightforward, \cite{meng2021internal} proposes an alternative that is equivalent to Eq.~\ref{eq:bayes} as follows:
\begin{equation}
    \hat{W} = \arg\max_{W}[\log p_{\theta}(W|X) -\log p_{\theta}^{\text{ILM}}(W) + \log p_{\text{LM}}(W) ] \label{eq:iml01}
\end{equation}
where $p_{\theta}^{\text{ILM}}(W)$ is provided by the so-called ``ILM" that is implicitly learned by training transcripts. In practice, Eq.~\ref{eq:iml01} is realized as follows:
\begin{equation}
    \hat{W} = \arg\max_{W}[\log p_{\theta}(W|X) -\lambda^{\text{ILM}} \log p_{\theta}^{\text{ILM}}(W) + \lambda \log p_{\text{LM}}(W) ] \label{eq:iml-lmf}
\end{equation}
where $\lambda^{\text{ILM}}$ and $\lambda$ are scaling factors for ILM and external LM respectively.
Now, to realize an inference process that is in harmony with Bayes theory under E2E ASR framework using external LM fusion, our focus is on how to estimate such an internal language model under the existing E2E framework with training transcripts. 
Besides, different E2E architectures has different ILME methods~\cite{meng2021internal}. In this paper, we employ two kinds of ASR frameworks, namely, Recurrent Neural Network Transducer~(RNN-T), as well as Listen and Spell (LAS) network that belongs to the attention-based encoder-decoder (AED) category.
\subsubsection{ILME for RNN-T}~\label{subsub:imle-rnn-t}
To be simple, RNN-T can be divided into encoder and decoder parts, and the decoder can be further divided into predict network and joint network for final recognition output. Let us denote the encoder output as $h^{\text{enc}}$, and the predict network output as $h^{\text{pred}}$, while the joint output is $z^{\text{joint}}$, we have
\begin{equation}
   z^{\text{joint}} =  \text{Joint}(h^{\text{enc}}, h^{\text{pred}})~\label{eq:rnn-t}
\end{equation}
To realize ILME for RNN-T, $z^{\text{joint}}$ is replaced to be independent of the encoder output $h^{\text{enc}}$.
\subsubsection{ILME for LAS}~\label{subsub:imle-las}
For LAS, with encoder output $h^{\text{enc}}$, decoder previous output $y_{u-1}$ as well as present hidden state $h_u^{\text{dec}}$, the decoder proceeds as follows:
\begin{align}
    a_u &= \text{MHA}(a_{u-1}, h^{\text{enc}}, h_u^{\text{dec}}) \label{eq:attend} \\
    c_u & = \sum_{t=1}^Ta_{u,t}h_t^{\text{enc}} \label{eq:scale} \\
    h_u^{\text{dec}} &= \text{DecoderRNN}(h_{u-1}^{\text{dec}}, \text{Concat}(\Tilde{y}_{u-1}, c_{u-1})) \label{eq:las-h}
\end{align}
where MHA refers to multi-head attention operation, $\Tilde{y}_{u-1}$ is  the embedding of decoder previous output.
Likewise, to perform ILME, we let $c_u$ be independent of encoder output $h^\text{enc}$ such that the decoder acts as an LM, i.e., the ILM here.

\subsection{ILME-based adaptive domain adaptation} ~\label{sub:almf}
ILME-based LM fusion has demonstrated significant performance improvement in both intra-domain (general domain) and cross-domain (target domain) ASR tasks when more external text data is available~\cite{meng2021internal, zeyer2021librispeech, zeineldeen2021investigating, liu2022internal}.

However, it is not Eq.~\ref{eq:iml-lmf} but Eq.~\ref{eq:asr-infer} that is  appropriate for general domain speech recognition. This is especially true when the source ASR model is trained with big data. 
More importantly, when we apply Eq.~\ref{eq:iml-lmf} to infer during recognition, the ASR system can only perform well on the specific target domain
utterance, and when the utterances from other domains are coming up, it is at risk of worse performance on the general domain.

\begin{table*}
\small
\begin{center}
\begin{tabular}{l|l||c|c|c|c||c|c|c|c}
  \hline
  \multicolumn{2}{c||}{} & \multicolumn{4}{c||}{CER (\%): RNN-T} & \multicolumn{4}{c}{ CER (\%): LAS} \\
  \cline{3-10}
\multicolumn{2}{c||}{} &\multicolumn{2}{c|}{Target: Search} & \multicolumn{2}{c||}{Target: Medical}&\multicolumn{2}{c|}{ Target: Search} & \multicolumn{2}{c}{ Target: Medical} \\
    \cline{3-10}
    \multicolumn{2}{c||}{} & Target & General & Target& General& Target & General & Target & General \\
    \hline
    \hline
   \multicolumn{2}{l||}{Baseline (no fusion)} & 21.88 & 13.89 & 4.47 & 13.89 & 21.37 & 8.99 & 4.70 & 8.99 \\
   \hline
  \hline
 \multirow{3}*{NNLM} & SF & 14.89 & 28.55 & 3.43 & 14.56 & 14.92 & 15.17 & 3.43 & 12.18 \\
  & ILME & 10.47 & 20.36 & 3.53 & 14.56 & 11.48 & 25.00 & 3.06 & 10.50  \\
  & ILME-ADA & 13.35 & 14.81 & 3.53 & 14.25 & 13.61 & 9.56 & 2.91 & 9.51 \\
  \hline
 \multirow{3}*{N-gram} & SF & 15.08 & 17.49 & 3.62 & 15.56 & 16.79 & 39.68 & 3.61 & 32.79 \\
  & ILME & 12.69 & 19.97 & 3.47 & 15.82 & 14.82 & 28.28 & 3.30 & 20.80  \\
  & ILME-ADA & 11.72 & 15.35 & 3.44 & 14.03 & 14.76 & 9.21 & 2.94 & 9.18 \\
  \hline
\end{tabular}
\end{center}
\caption{CERs (\%) of both RNN-T and LAS models on the voice search  (\texttt{Search}) and  medical domain (\texttt{Medical}) test sets with the proposed ILME-ADA method. For each domain, an external NNLM and n-gram LM are trained with \textbf{ONLY} target domain text data. All results are obtained with both $\lambda$ and $\lambda^{\text{ILM}}$ being tuned to the best on the corresponding target domains.}
\label{tab:res-ilme-ada}
\end{table*}

Based on the above-mentioned consideration, in this paper, we propose an ILME-based adaptive LM fusion method.
Specifically, we want our LM fusion method not only to perform well on the target domain but its performance on the general domain would also be minimally compromised simultaneously. In other words, given an utterance without prior knowledge of whether it is from the general domain or it is from the target domain, the proposed LM fusion method can perfectly handle it. Simply put, the proposed method is formulated as:
\begin{multline}
       \hat{W} = \arg\max_{W}[\log p_{\theta}(W|X) -\lambda^{\text{ILM}} \log p_{\theta}^{\text{ILM}}(W) + \\ \max(\lambda^{\text{ILM}} \log p_{\theta}^{\text{ILM}}(W), \lambda \log p_{\text{LM}}(W)) ] \label{eq:imle-adapt}
\end{multline}
Eq.~\ref{eq:imle-adapt} actually implies an adaptive LM fusion. This is because for each decoding step, we compare the log likelihood scores between ILM and external LM, and the maximum score is selected. For instance, if $p_\theta^{\text{ILM}}(W) > p_{\text{LM}}(W)$ we employ Eq.~\ref{eq:asr-infer} to infer, otherwise, we employ Eq.~\ref{eq:iml-lmf} to perform inference. 
In what follows, we are demonstrating its efficacy for domain adaptation.

\section{Experiments}
\subsection{Datasets}

The RNN-T and LAS models are trained on ASR Mandarin training sets containing 200k and 100k hours of speech respectively.
The speech data is extracted from short videos that are anonymized before transcript labeling.
Since both training sets are big, it is sensible for us to think of them as general sets. Moreover, to define a general test set, we select a disjoint subset that has the same source as the training set as our general test set which contains 3751 utterances.
To evaluate domain adaptation performance, we define two test sets from two domains respectively. One is for  book inquiry, denoted as \texttt{Search} test set, and the other is for the doctor and patient conversation in medical domain, denoted as \texttt{Medical} accordingly. Table~\ref{tab:da-data} presents the details of the data descriptions for the domain adaptation task. We train 2 pairs of LMs with the training data, one is the NNLM, and another is n-gram LMs with n=5. 

\begin{table}[h]
    \centering
    \begin{tabular}{c|c|cc}
    \hline
     \multirow{2}{*}{Target Domain }    & Train & \multicolumn{2}{c}{Test} \\
     \cline{2-4}
     & characters (M) & character (K) & hours \\ 
     \hline
      Search   &  412  &33.03 & 3.25\\
      Medical  &  519 &287.05 & 22.83\\
      \hline
    \end{tabular}
    \caption{Data description for the target domain adaptation evaluation.}
    \label{tab:da-data}
\end{table}

\subsection{Models}
\subsubsection{ASR Models}
For RNN-T ASR modeling, the encoder of the RNN-T model contains 30 layers of DFSMN \cite{zhang2018deep}, each with 512 memory cells, and 2048 hidden units. The prediction network consists of a 2-layer LSTM with 2048 hidden size. The joint network consists of a linear layer followed by the ReLU function with the hidden size being 768. 

For LAS ASR modeling,
the encoder of the LAS model is a 18-layer Transformer. Each attention layer outputs a 512 dimension vector with 8 heads. The intermediate fully-connected layer has a dimension of 2048 with the dropout rate of 0.1. The decoder adopts a 4-layer LSTM with 1024 hidden units. 
The model is first trained using the cross-entropy loss and then  the minimum word error(MWER) criterion \cite{prabhavalkar2018minimum} training is practiced. 

\subsubsection{Language Models}

We train NNLM with two LSTM-Projection \cite{sak2014long} layers, each comprising 2048 hidden LSTM units followed by a linear layer mapping to a 512 dimension vector. We use the Adam optimizer \cite{kingma2015adam} with a maximum learning rate of 1e-3 and adopt early stopping based on the performance on the development sets. For the WFST fusion, 5-gram LMs with Kneser-Ney smoothing \cite{james2000modified} are first trained with the KenLM tookit \cite{heafield2011kenlm} and then converted using OpenFST \cite{allauzen2007openfst}. 

To compare different fusion methods,
we tune 
all LM scales $\lambda$ and $\lambda^{\text{ILM}}$  with grid search that falls in (0.0, 1.0], as well as search for the best blank weighting factor for RNN-T  and other parameters involved with both ASR frameworks for best inference results. For RNN-T models, we adopt the zero ILME method \cite{meng2021internal} while for LAS models we adopt the sequence-level encoder average ILME method~\cite{zeineldeen2021investigating}.  We also compare the results of employing different underlying ILME methods for the analysis experiments. 

\begin{table}
    \centering
    \begin{tabular}{cc||c| c|c|c}
    \hline
     \multicolumn{2}{c||}{\multirow{2}{*}{Model}} & \multicolumn{2}{c|}{PPL (Search)} &
     \multicolumn{2}{c}{PPL (Medical)} \\
     \cline{3-6} 
     & &Target & General & Target & General \\
     \hline
     \hline
     \multirow{2}*{LM} & NNLM &12.15 & 169.05 &16.58  &100.32  \\
     & n-gram LM & 34.47 & 239.86 & 24.61 & 148.13 \\
     \hline
    \multirow{2}*{ILM} & RNN-T &460.76 & 132.86 & 186.31 & 132.86  \\
     & LAS &395.61 &110.96 &299.55 & 110.96 \\
     \hline
    \end{tabular}
    \caption{Perplexity results on both general and target domain test sets calculated with the external  LMs and ILMs respectively.}
    \label{tab:ppl}
    \vspace{-5mm}
\end{table}

\subsection{Results}~\label{sub:res}
\subsubsection{Results of perplexity evaluation}
Table~\ref{tab:ppl} presents the perplexity results on both general and target domain test sets with diverse language modeling methods.
From Table~\ref{tab:ppl}, we see that both NNLMs and n-gram external LMs achieve much lower perplexity results on the target domains while  higher ones on the general domain test sets. This means the target domain data are generally different from the general domain. 
Besides, the n-gram LM perplexity is much higher than 
the NNLMs' on either domain. 
For ILM, the perplexities on the general test set are much lower than those on the corresponding target test sets. This also reflects the difference between the general and domain specific data sets.
Interestingly, LAS based ILME gets better perplexity  on the general test set, while it's better on the \texttt{Search} domain test set and worse on the \texttt{Medical} domain test set. 
However for ILM, we are only interested in its performance on the general test set, and lower perplexity means a better ILM estimation method. Probably, 
the encoder output average-based ILME  method is better than the zero-out ones. We will give a further analysis of this in Section~\ref{subsub:analysis}.
Last but not the least,
the NNLM trained with target domain data has obtained even better perplexity than either ILME's on the general test set, 100.32 versus 132.86 (RNN-T), and 100.32 versus 110.96 (LAS) respectively. This might suggest our target domain data contains some  sentences from the general domain that is learned by the NNLM. The observation needs to be further clarified in the future. Moreover, the lower perplexity from the external NNLM on the general test set might affect the efficacy of the proposed method from Equation~\ref{eq:imle-adapt}, particularly on the general test set.



\subsubsection{Results of the proposed ILME-ADA method}~\label{subsub:res-ilme-ada}
Table~\ref{tab:res-ilme-ada} reports the CER results of the proposed ILME-ADA method for domain adaptation, compared with shallow fusion (SF), as well as ILME fusion methods.
What Table~\ref{tab:res-ilme-ada} reveals is a little bit surprising. the n-gram LM based ILME-ADA
gets  better performance than the NNLM except for the RNN-T case on the general domain, 15.35 versus 14.81, and for the LAS case on the \texttt{Medical} domain, 2.94 versus 2.91. This is different from what has been observed for shallow fusion, as well as  ILME fusion, where the NNLM has shown obvious edge  over the n-gram LM under different scenarios.
This suggests the stronger external NNLM is not decisive for bringing better performance, but the capability of capturing different knowledge between the ILM and the corresponding external LM is crucial for the proposed ILME-ADA method for domain adaptation. 

For the n-gram LM as external LM on the target domain, the proposed ILME-ADA method has achieved 22.3\% (\texttt{Search}), 4.97\% (\texttt{Medial}) CER reduction (CERR) for RNN-T streaming models, while 12.1\% (\texttt{Search}), and 18.6\% (\texttt{Medical}) CERR for LAS compared with SF method respectively. Compared with the ILME, the proposed method also has better performance.
On the general domain, the proposed method drops 10.5\% (\texttt{Search}), 1.0\% (\texttt{Medical})
relative CER  for the RNN-T, and 2.4\% (\texttt{Search}),  2.1\% (\texttt{Medical}) relative CER for the LAS. All are significantly better than what SF and ILME could have achieved.  We note that all results are obtained with both $\lambda$ and $\lambda^{\text{ILM}}$ being tuned to the best. 

While for the NNLM as ELM on the target domain, Although Table~\ref{tab:res-ilme-ada} reveals the advantage of the proposed method over the conventional SF method generally. It has yielded worse results compared with corresponding ILME method. For the NNLM, we note that its perplexity on the general domain test set is even better than what the ILMs can obtain. This indicates the proposed method has lost its impact in most cases during inference process from Equation~\ref{eq:imle-adapt}, as a result, yielding worse results. Further analysis of the underlying reason will be left for future work.

\subsubsection{Different ILME method comparison}~\label{subsub:analysis}

\begin{table}
\small
\begin{center}
\begin{tabular}{l||c|c|c}
  \hline
     & $\lambda/\lambda^{\text{ILM}}$ & CER(\%) & CERR \\
    \hline
      \hline
    Baseline & 0.0/0.0 & 4.70 & 0.0 \\
  \hline
  \hline
  ILME (AvgH) & 0.2/0.3 &  3.46 & 26.4 \\
  ILME-ADA (AvgH) & 0.6/0.6  & 2.95 & 37.3 \\
  \hline
  ILME (Zero) & N.A. & - & - \\
  ILME-ADA (Zero) & 0.3/0.1 & 4.18 & 11.1 \\
  \hline
  ILME (OTCL) & 0.1/0.1 & 3.74 & 20.5 \\
  ILME-ADA (OTCL) & 0.4/0.3  & 3.27 & 30.5 \\
  \hline
  ILME (LSCL) & 0.1/0.1  & 3.73 & 20.7 \\
  ILME-ADA (LSCL) & 0.5/0.5 & \textbf{2.90} & 38.3 \\
  \hline
\end{tabular}

\caption{CERs (\%) of the LAS model on the \texttt{Medical} test set with different ILM estimation methods, where external LM is NNLM. All results are obtained with no more than a relative 3.0\% CER degradation being allowed on the general domain test set.}
\label{tab:ilm}
\end{center}
\vspace{-3mm}
\end{table}

The proposed ILME-ADA method has shown a clear advantage for domain adaptation. However, it is closely
related to how the ILM is estimated. 
In what follows, we show the efficacy of  different ILME  methods using the LAS ASR modeling framework on the  \texttt{Medical} test set, where NNLM is employed as external LM. 
Concretely, the ILM is estimated with methods including average encoder output sequence (AvgH)~\cite{zeineldeen2021investigating}, zeroing-out context vector (Zero)~\cite{meng2021internal}, one-time context vector learning (OTCL)~\cite{liu2022internal}, as well as label-synchronous context vector learning (LSCL)~\cite{liu2022internal} respectively. Table~\ref{tab:ilm} reports CER results  of which less than 3.0\%  performance degradation is allowed on the general domain. 
From Table~\ref{tab:ilm}, we can observe both ILME with ``AvgH" and ``LSCL" methods earn big margin CERRs over the ``Zero" method. For detailed analysis, one can refer to \cite{liu2022internal} for more details. Besides, Table~\ref{tab:ilm} also indicates the ``Zero" method cannot achieve less than 3.0\% CERR on the general domain test set.


\section{Conclusions}
In this paper, we proposed an ILME-based adaptive language model fusion method for domain adaptation. When only the target domain text data is available to build external LMs, the proposed method  can achieve obvious performance improvement over both shallow  and ILME fusions for domain adaptation
 while the performance on the general domain test set (source domain data) is minimally influenced, $\sim$3.0\% CER drop relatively. We found when NNLM and n-gram LM are employed as external LM, the advantage of the proposed method  is particularly remarkable when using the latter.
 Additionally, we also compared the efficacy of the proposed method with different internal language model estimation methods, and found the zeroing-out encoder output to estimate ILM for the transformer encoder is not desirable.

\section{acknowledgement}
Part of the work has been done by our prior colleagues Yongbin You and Xuezhi Wang. Many thanks should be given to their pioneering contributions.
\vfill\pagebreak
\clearpage
\bibliographystyle{ieeetr}
\bibliography{haihua}

\begin{thebibliography}{10}

\bibitem{graves2014towards}
A.~Graves and N.~Jaitly, ``Towards end-to-end speech recognition with recurrent
  neural networks,'' in {\em International conference on machine learning},
  pp.~1764--1772, PMLR, 2014.

\bibitem{chan2016listen}
W.~Chan, N.~Jaitly, Q.~Le, and O.~Vinyals, ``Listen, attend and spell: A neural
  network for large vocabulary conversational speech recognition,'' in {\em
  2016 IEEE international conference on acoustics, speech and signal processing
  (ICASSP)}, pp.~4960--4964, IEEE, 2016.

\bibitem{vaswani2017attention}
A.~Vaswani, N.~Shazeer, {\em et~al.}, ``Attention is all you need,'' {\em
  Advances in NIPS}, vol.~30, 2017.

\bibitem{chiu2018state}
C.-C. Chiu, T.~N. Sainath, Y.~Wu, R.~Prabhavalkar, P.~Nguyen, Z.~Chen,
  A.~Kannan, R.~J. Weiss, K.~Rao, E.~Gonina, {\em et~al.}, ``State-of-the-art
  speech recognition with sequence-to-sequence models,'' in {\em 2018 IEEE
  International Conference on Acoustics, Speech and Signal Processing
  (ICASSP)}, pp.~4774--4778, IEEE, 2018.

\bibitem{gulati2020conformer}
A.~Gulati, J.~Qin, {\em et~al.}, ``Conformer: Convolution-augmented transformer
  for speech recognition,'' {\em arXiv:2005.08100}, 2020.

\bibitem{hoy2018alexa}
M.~B. Hoy, ``Alexa, siri, cortana, and more: an introduction to voice
  assistants,'' {\em Medical reference services quarterly}, vol.~37, no.~1,
  pp.~81--88, 2018.

\bibitem{sainath2020streaming}
T.~N. Sainath, Y.~He, B.~Li, {\em et~al.}, ``A streaming on-device end-to-end
  model surpassing server-side conventional model quality and latency,'' in
  {\em Proc. of ICASSP}, IEEE, 2020.

\bibitem{li2022language}
B.~Li, T.~N. Sainath, {\em et~al.}, ``A language agnostic multilingual
  streaming on-device asr system,'' {\em arXiv:2208.13916}, 2022.

\bibitem{bell2020adaptation}
P.~Bell, J.~Fainberg, O.~Klejch, J.~Li, S.~Renals, and P.~Swietojanski,
  ``Adaptation algorithms for neural network-based speech recognition: An
  overview,'' {\em IEEE Open Journal of Signal Processing}, vol.~2, pp.~33--66,
  2020.

\bibitem{rosenberg2019speech}
A.~Rosenberg, Y.~Zhang, B.~Ramabhadran, Y.~Jia, P.~Moreno, Y.~Wu, and Z.~Wu,
  ``Speech recognition with augmented synthesized speech,'' in {\em 2019 IEEE
  automatic speech recognition and understanding workshop (ASRU)},
  pp.~996--1002, IEEE, 2019.

\bibitem{joshi2022simple}
R.~Joshi and A.~Singh, ``A simple baseline for domain adaptation in end to end
  asr systems using synthetic data,'' in {\em Proceedings of The Fifth Workshop
  on e-Commerce and NLP (ECNLP 5)}, pp.~244--249, 2022.

\bibitem{toshniwal2018comparison}
S.~Toshniwal, A.~Kannan, C.-C. Chiu, Y.~Wu, T.~N. Sainath, and K.~Livescu, ``A
  comparison of techniques for language model integration in encoder-decoder
  speech recognition,'' in {\em 2018 IEEE spoken language technology workshop
  (SLT)}, pp.~369--375, IEEE, 2018.

\bibitem{sriram2018cold}
A.~Sriram, H.~Jun, S.~Satheesh, and A.~Coates, ``Cold fusion: Training seq2seq
  models together with language models,'' {\em Proc. Interspeech 2018},
  pp.~387--391, 2018.

\bibitem{gulcehre2015using}
C.~Gulcehre, O.~Firat, K.~Xu, K.~Cho, L.~Barrault, H.-C. Lin, F.~Bougares,
  H.~Schwenk, and Y.~Bengio, ``On using monolingual corpora in neural machine
  translation,'' {\em arXiv preprint arXiv:1503.03535}, 2015.

\bibitem{variani2020hybrid}
E.~Variani, D.~Rybach, C.~Allauzen, and M.~Riley, ``Hybrid autoregressive
  transducer (\text{HAT}),'' in {\em ICASSP 2020-2020 IEEE International
  Conference on Acoustics, Speech and Signal Processing (ICASSP)},
  pp.~6139--6143, IEEE, 2020.

\bibitem{mcdermott2019density}
E.~McDermott, H.~Sak, and E.~Variani, ``A density ratio approach to language
  model fusion in end-to-end automatic speech recognition,'' in {\em 2019 IEEE
  Automatic Speech Recognition and Understanding Workshop (ASRU)},
  pp.~434--441, IEEE, 2019.

\bibitem{meng2021internal}
Z.~Meng, S.~Parthasarathy, E.~Sun, Y.~Gaur, N.~Kanda, L.~Lu, X.~Chen, R.~Zhao,
  J.~Li, and Y.~Gong, ``Internal language model estimation for domain-adaptive
  end-to-end speech recognition,'' in {\em 2021 IEEE Spoken Language Technology
  Workshop (SLT)}, pp.~243--250, IEEE, 2021.

\bibitem{zeyer2021librispeech}
A.~Zeyer, A.~Merboldt, W.~Michel, R.~Schl{\"u}ter, and H.~Ney, ``Librispeech
  transducer model with internal language model prior correction,'' {\em arXiv
  preprint arXiv:2104.03006}, 2021.

\bibitem{zeineldeen2021investigating}
M.~Zeineldeen, A.~Glushko, W.~Michel, A.~Zeyer, R.~Schl{\"u}ter, and H.~Ney,
  ``Investigating methods to improve language model integration for
  attention-based encoder-decoder \text{ASR} models,'' {\em arXiv preprint
  arXiv:2104.05544}, 2021.

\bibitem{liu2022internal}
Y.~Liu, R.~Ma, H.~Xu, Y.~He, Z.~Ma, and W.~Zhang, ``Internal language model
  estimation through explicit context vector learning for attention-based
  encoder-decoder \text{ASR},'' {\em arXiv preprint arXiv:2201.11627}, 2022.

\bibitem{choudhury2022likelihood}
C.~Choudhury, A.~Gandhe, X.~Ding, and I.~Bulyko, ``A likelihood ratio based
  domain adaptation method for \text{E2E} models,'' {\em arXiv preprint
  arXiv:2201.03655}, 2022.

\bibitem{zhang2018deep}
S.~Zhang, M.~Lei, Z.~Yan, and L.~Dai, ``\text{Deep-FSMN} for large vocabulary
  continuous speech recognition,'' in {\em 2018 IEEE International Conference
  on Acoustics, Speech and Signal Processing (ICASSP)}, pp.~5869--5873, IEEE,
  2018.

\bibitem{prabhavalkar2018minimum}
R.~Prabhavalkar, T.~N. Sainath, Y.~Wu, P.~Nguyen, Z.~Chen, C.-C. Chiu, and
  A.~Kannan, ``Minimum word error rate training for attention-based
  sequence-to-sequence models,'' in {\em 2018 IEEE International Conference on
  Acoustics, Speech and Signal Processing (ICASSP)}, pp.~4839--4843, IEEE,
  2018.

\bibitem{sak2014long}
H.~Sak, A.~Senior, and F.~Beaufays, ``Long short-term memory based recurrent
  neural network architectures for large vocabulary speech recognition,'' {\em
  arXiv preprint arXiv:1402.1128}, 2014.

\bibitem{kingma2015adam}
D.~P. Kingma and J.~Ba, ``Adam: A method for stochastic optimization,'' in {\em
  ICLR (Poster)}, 2015.

\bibitem{james2000modified}
F.~James, ``Modified kneser-ney smoothing of n-gram models,'' {\em Research
  Institute for Advanced Computer Science, Tech. Rep. 00.07}, 2000.

\bibitem{heafield2011kenlm}
K.~Heafield, ``\text{KenLM}: Faster and smaller language model queries,'' in
  {\em Proceedings of the sixth workshop on statistical machine translation},
  pp.~187--197, 2011.

\bibitem{allauzen2007openfst}
C.~Allauzen, M.~Riley, J.~Schalkwyk, W.~Skut, and M.~Mohri, ``\text{OpenFst}: A
  general and efficient weighted finite-state transducer library,'' in {\em
  International Conference on Implementation and Application of Automata},
  pp.~11--23, Springer, 2007.

\end{thebibliography}

\end{document}